# Machine Learning and Ensemble Approach Onto Predicting Heart Disease

Aaditya Surya
*Computer Science*
*Lincoln School*
Kathmandu Nepal
aaditya.surya0@gmail.com

*Abstract*— The four essential chambers of one's heart that lie in the thoracic cavity are crucial for one's survival, yet ironically prove to be the most vulnerable. Cardiovascular disease (CVD) also commonly referred to as heart disease has steadily grown to the leading cause of death amongst humans over the past few decades. Taking this concerning statistic into consideration, it is evident that patients suffering from CVDs need quick and correct diagnosis in order to facilitate early treatment to lessen the chances of fatality. This paper attempts to utilize the data provided to train classification models such as Logistic Regression, K Nearest Neighbors, Support Vector Machine, Decision Tree, Gaussian Naive Bayes, Random Forest, and Multi-Layer Perceptron (Artificial Neural Network) and eventually using a soft voting ensemble technique in order to attain as many correct diagnoses as possible.

*Keywords—cardiovascular disease, classification, ensemble learning,*

## I. Introduction

Cardiovascular diseases (commonly known as CVDs) are disorders pertaining to the function of the heart, and prove to be the leading cause of natural death across the globe. The WHO (World Health Organization) listed cardiovascular disease in 2019 to account for 32% of all global deaths while further adding that most cardiovascular diseases were largely preventable through addressing behavioral risk factors (such as unhealthy diets, obesity, tobacco use, alcohol consumption, etc.)[1]. Taking into consideration the preventability of cardiovascular deaths, a more concerning statistic shows the severe mortality rates that heart disease holds. In countries of high development such as the United States the mortality reaches almost 1 in 4 people, while in more developing countries such as within the MENA (Middle East and North Africa) region the mortality rates soar even higher to around just under 40%. The way to tackle this high mortality rate is to ensure early diagnosis in order to facilitate a quick and correct treatment process for patients. The term Cardiovascular disease/Heart disease encompasses a wide array of deformities within the heart. Table 1 below displays the various types of heart disease.

[1] https://www.who.int/health-topics/cardiovascular-diseases

Table.1. Types of Cardiovascular Disease and Their Descriptions

| ID | Type of Cardiovascular Disease | Description |
|----|-------------------------------|-------------|
| 1. | Arrhythmia | An arrhythmia is an abnormality within the heart where there is an irregular or abnormal heartbeat. Oftentimes these situations are not too significant, however, further complications can make arrhythmia a serious or even fatal disease. |
| 2. | Cardiac arrest | Cardiac arrest is essentially when a person's heart fails to keep beating. Due to this, blood flow across the body comes to a halt and essential organs such as the lungs and the brain do not receive the required amount of blood. |
| 3. | Congestive heart failure | Congestive heart failure is quite a serious medical condition in which the heart is not able to pump blood efficiently across the body. |
| 4. | Congenital heart disease | Congenital heart disease is often a birth defect as it occurs since birth. The disease causes a defect that may change the way blood flows through one's heart. |
| 5. | Coronary artery disease | Coronary heart disease occurs when an artery has been blocked by the buildup of a fatty substance. This buildup restricts the flow of blood and may cause angina, or chest pain and discomfort. |
| 6. | High Blood Pressure | High blood pressure also commonly referred to as hypertension is a condition where the pressure of blood against one's blood vessels is consistently too large. |
| 7. | Peripheral artery disease | Peripheral artery disease is essentially the shrinking or narrowing of the arteries that take blood away from the heart, often to other organs or parts of the body. |
| 8. | Stroke | A stroke happens when blood supply to your brain has reduced or has been stopped. This stops your blood from getting essential nutrients and the oxygen needed to function and hence causes brain cells to begin dying. |

With the development of machine learning technology and the further improved accessibility of heart disease data from sources such as the University of California-Irvine (UCI) Machine Learning Data Repository, the feasibility of employing machine learning models to obtain a prediction has dramatically increased. This project primarily aims to employ 6 classification model techniques: Logistic Regression, K Nearest Neighbors, Support Vector Machine, Decision Tree, Gaussian Naive Bayes, Random Forest, and Multi-Layer Perceptrons (an Artificial Neural Network), and ultimately aims to ensemble these supervised techniques to attain a potential diagnosis.

This project allows for a machine to diagnose a patient who is likely to have heart disease through finding similarities between a potential patient's symptoms and a previous patient in the database. With this in mind, the project utilizes the heart disease dataset available on the UCI repository that provides a patient's medical history, some further diagnostic metrics (such as resting blood pressure, chest pain, etc.), and a target value that identifies the diagnosis of heart disease. By training this provided data under the 6 machine learning models provided above, the project hopes to attain the most robust diagnosis.

## II. Literature Review

Considering the severity of heart disease, and the prominence of predicting heart disease, quite a lot of work has been conducted within this field.

Kohli et al. [1] propounded through their works that through hyperparameter optimization on models such as Logistic Regression, KNN, SVM, Random Forest, and Decision Tree, a higher accuracy can be achieved with comparison to just the default models of each classification technique. Through keeping the kernel linear, gamma as auto and C=2, the researchers were able to obtain accuracies as high as 90.32% within their attempt at predictions of Abstract Heart-disease.

Vakili Meysam, et al. [2] lists multiple other research papers that have been analyzed for the sake of this project to conclude the best evaluation metrics for different algorithms. In section 3.3 of the research paper definitions and formulas for multiple classification performance evaluation metrics have been listed in order to show the benefits and detriments of each metric. The listed metrics are precision, recall, f1-score, accuracy, confusion matrix, and ROC-AUC scores. These metrics are primarily used to analyze how well a classification model has performed.

Gavhane et al. [3] through their work proved that "automated medical diagnosis and prediction systems would prove extremely favorable" when attempting to prevent cardiovascular deaths. The paper further enforced the unsatisfactory results and medical costs patients face through their diagnosis process, and how an automated diagnosis would prove to be much more favorable.

Chao-Ying Joanne Peng et al. [4] discusses the fundamentals of a logistic regression model for classification prediction. The research paper mentions how the logistic regression model is based under the regression of a logit (natural logarithm of an odds ratio) function. This creates a sigmoid function allowing for one to make a graph with a dichotomous outcome variable while maintaining a continuous predictor.

Himanshu Sharma et al. [5] proved that the emergence of deep learning and artificial intelligence show optimistic results in fields such as medical diagnosis and that this is "an open domain waiting to be implemented in heart disease prediction.

Phillip Probst et al. [6] addresses the importance of hyperparameter optimization when dealing with machine learning algorithms. The paper suggests that hyperparameter optimization is performed through a tuning procedure in which parameters of certain models are optimized to suit the problem at hand. The main question this paper tackles is which hyperparameters should be tuned, how they should be tuned, and in which ranges should this tuning occur. The paper shows the general notation when used in hyperparameter optimization, the measures to which one can optimize hyperparameters, and further how one can measure the optimization that has occurred.

Senthil Kumar et al. [7] implemented a model with multiple classification techniques such as Decision Trees, Random Forest, Genetic algorithms, and Naive Bayes to attain high accuracies. Using Hybrid Random Forest with Linear Model (HRFM) the researchers obtained an astounding accuracy level of 88.7%.

J. Nageswara Rao et al. [8] further developed an ENDDP (Enhanced New Dynamic Data Processing) Algorithm in order to attempt to predict the early stages of heart disease.

Akinsola, et al. [9] provide an overview of classification and comparison models that can be used within machine learning problems. The paper targets only supervised machine learning algorithms which basically represent algorithms that input externally supplied data to produce general classification hypotheses that can be used to make predictions. The paper lists out the multiple classification algorithms and how they work, these classifiers include linear classifiers, logistic regression, support vector machines, k-means, Naive Bayesian networks, decision trees, multilayer perceptrons, and many more. The paper also lists out the process that is undergone during supervised machine learning which in summary is identifying a problem, identifying required data, data pre-processing, defining the training set, algorithm selection, training, evaluation with a test set, parameter tuning, retraining, and evaluation, and measuring the result analysis of the model.

Nikhil Kumar et al. [10] through working with algorithms such as Logistic Model Tree, Random Forest, Naive Bayes, KNN, SVM and more to predict heart disease proved through their paper that accuracy levels are positively associated with the number of attributes used.

Niklas Lavesson et al. [11] summarize the methods present in a majority voting ensemble and provide a comparative analysis of these voting schemes in the context of malware detection. The paper details how ML can be broken into supervised learning algorithms to generate a classifier or regression function, and how in order to achieve maximum accuracy it is often best to create an ensemble-based majority voting platform. Ensemble in the paper is described as the capability of combining multiple models for improved accuracy as ensembles perform better than a single model due to the "diversity of base models." It further details how voting in machine learning should include specific characteristics to ensure that there is fair voting. In order to achieve this, the article also provides a useful diagram showing the process required in ensemble voting.

## III. DATA SOURCE

A. For the purpose of this project, the University of California-Irvine Machine Learning Repository dataset on heart disease shall be used. The UCI Machine Learning Repository is a well-known repository that is used frequently by members of the machine learning community for the purpose of analysis on algorithms. The specific dataset that shall be used was donated through the efforts of the Hungarian Institute of Cardiology, University Hospital, Zurich, University Hospital, Basel, and V.A. Medical Center in Long Beach. The database itself contains 75 attributes (not including the target value that determines the diagnosis) with 303 instances for each attribute. For the purpose of this experiment, however, a subset of 14 of those attributes shall be used. The dataset does contain missing/null values, however, through data evaluation, it was determined that the 14 attributes that are being used do not contain any null values. Through providing a varied demographic of patients through different age groups and genders the dataset, though small, helps correctly sample the population of heart disease patients well. Furthermore, through giving access to medical attributes such as resting blood pressure, cholesterol levels, resting electrocardiographic results, etc. the dataset facilitates employing machine learning models to predict if a patient is diagnosed with heart disease or not. A description of the attributes that will be used for means of prediction is displayed below in 'Table 2.'

Table.2. Descriptions of the Attributes Used for Prediction

| ID | Observation | Description | Values |
|---|---|---|---|
| 1. | Age | Age of patient in years | Continuous |
| 2 | Sex | Sex of patient | Male/Female |
| 3. | CP | Chest pain | Value 1: typical angina Value 2: atypical angina Value 3: non-anginal pain Value 4: asymptomatic |
| 4. | Trestbps | Resting blood pressure | Continuous |
| 5. | Chol | Serum cholesterol in mg/dl | Continuous |
| 6 | FBS | Fasting blood sugar >120 mg/dl | <, or > 120 mg/dl |
| 7. | Restecg | Resting electrocardiographic results | Value 0: normal Value 1: having ST-T wave abnormality (T wave inversions and/or ST elevation or depression of > 0.05 mV) Value 2: showing probable or definite left ventricular hypertrophy by Estes' criteria |
| 8. | Thalach | Maximum heart rate achieved | Continuous |
| 9. | Exang | Exercise induced angina | Yes/No |
| 10. | Oldpeak | ST depression induced by exercise relative to rest | Continuous |
| 11. | Slope | The slope of the peak exercise ST segment | Value 1: upsloping Value 2: flat Value 3: downsloping |
| 12. | Ca | Number of major vessels (0-3) colored by fluoroscopy | 0-3 |
| 13. | Thal | Defect type | 3 = normal 6 = fixed defect 7 = reversible defect |
| 14. | Target | Diagnosis of heart disease | 0 = Healthy 1 = Defect |

## IV. PROPOSED METHODOLOGY

The methodology that will be used within this project will follow the model displayed in Figure 1 displayed below:

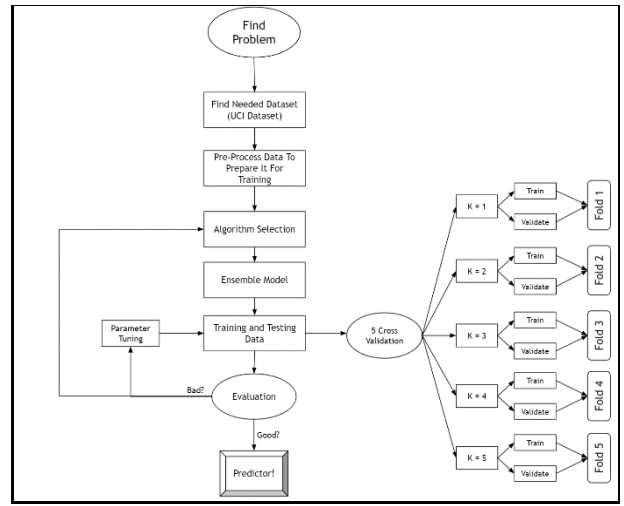

Fig. 1 . General Overview of the Methodology

### A. Identifying a Problem

The initial stage of the methodology is to introduce and identify the problem at hand. The problem within this project is rather clear in attempting to get a consistently correct prediction of heart disease using machine learning.

### B. Obtaining a Dataset

The secondary stage is to find a reliable dataset. Within this scenario, the University of California-Irvine has an extensive machine learning data repository, which as a matter of fact, also happens to contain a heart disease dataset. One needs to analyze this dataset and see which attributes are going to be used. For this experiment all 14 attributes of the dataset were utilized in the prediction.

### C. Data Analysis and Data Preprocessing

For efficient data engineering a thorough process of data analysis and preprocessing must be undertaken. The primary stage of analysis consists of addressing whether or not the dataset contains any null or missing values. The dataset itself is shown to contain missing values, however, the 14 attributes being used for this project do not have any null values. This allows for one to move further with data analysis and consider any imbalances in the datasets. Figure 2 displayed below addresses the distribution of the target variable through the dataset. The graph displays that 45.5% of the instances are normal, and 54.5% of instances have a heart defect. This shows that the data is slightly imbalanced, however, this imbalance is quite negligible.

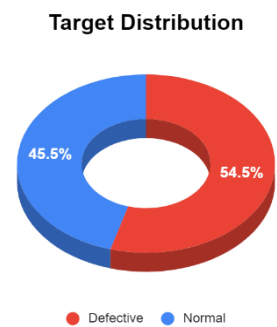

Fig. 2 . Target Distribution

With regards to the patients in the dataset, the distributions in genders are shown below in Figure 3. With this in mind, the dataset fails to consider non-binary individuals when addressing genders, and hence only considers males and females. This limitation to the dataset, however, should again prove to be rather negligible. The graph shows that there are 207 males, and 96 females in the dataset, showing a large imbalance towards males. This may be a limiting factor for the dataset.

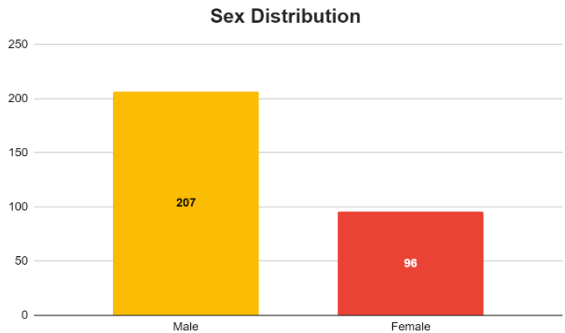

Fig. 3 . Sex Distribution

Considering both these imbalances, as the imbalance within the target values is negligible, and the gender gap, though large, should not have too much of an impact, synthetic sampling was not considered for this dataset.

The quantitative features within the dataset had uneven scaling which can cause a delay in the convergence of optimization-based classifiers. [11] Due to this standardization was performed on quantitative features in order to minimize the effect of this issue.

Lastly a correlation heat map was drawn in order to consider removing unnecessary attributes from the predictive model. This heatmap is shown as Figure 4 below. Through this heat map it was identified that fbs (Fasting Blood Sugar) had a noticeably low correlation with the target variable, however, it was chosen not to remove this attribute as the minimal correlation could be misleading considering another model that was non-linear could be utilized to approach fasting blood sugar.

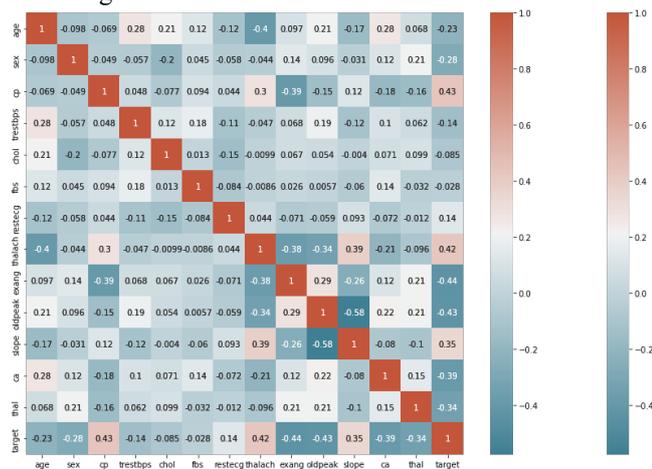

Fig. 4 . Correlation Heatmap

### D.  Algorithm Selection

This project will utilize seven different algorithms that are defined below:

*a) Logistic Regression:* The Logistic Regression model also commonly known as the logit model utilizes the following function in order to classify between two target responses:

$$\log\left(\frac{p}{1-p}\right)$$

This creates a sigmoid logistic model function in which the range of probability is always between 0 and 1. Then a threshold is set. When the probability of success is found to be over the predefined threshold the function predicts a target of 1, and similarly if the probability found is under the threshold the function predicts a target of 0. Most often the threshold is set at a value of 0.50.

*b) K-Nearest Neighbors:* In a K-Nearest Neighbors (KNN) algorithm a data point that has not been classified is plotted on a plain with training data. A constant "k" is defined which then allows for the algorithm to select a k amount of neighbors using the points with the smallest Euclidean distance from the unclassified chosen data point. The chosen data point is then classified based on the category the most amount of its neighbors have been classified into.

*c) Support Vector Machine:* A Support Vector Machine (SVM) classifies through generalising the data to find two different sets. These sets are placed on a hyperplane on which a maximum margin is drawn in order to differentiate between the positive and negative hyperplanes, hence helping one to classify each point as either positive or negative.

*d) Decision Tree:* Decision trees work by using algorithms to split a node into two or more sub-nodes, and to further repeat this split process until a final decision can be made. The larger number of sub nodes allows for an increase in the homogeneity in further sub nodes, hence in a sense increasing the clarity at which one can approach a final decision with respect to the target variable.

*e) Gaussian Naïve Bayes:* The principle of a gaussian naive bayes algorithm is that a GNB classifier calculates the z-score distance between the point and each class mean. This essentially means that there is a comparison between the probability that the point comes from the distribution of Class A (the Z-score from the class B mean and standard deviations), and the probability that the distribution comes from class B (the Z-score from the class B mean and standard deviations)

*f) Random Forest Classifier:* Random forest is a supervised learning algorithm that essentially ensembles multiple decision trees. The combination of multiple decision trees provides a better overall result.

*g) Multi-Layer Perceptrons Neural Network:* A multilayer perceptron model is a supervised deep learning technique which in principle works by establishing three layers: an input layer, a hidden layer, and an output layer. The input layer allows for the features to be inputted into the model, which is then pushed through the MLP by taking a dot product of the input with the weights that are assigned to each node in the hidden layer. Once these dot products have been calculated activation functions push the outputs of the first calculation to deeper layers within the hidden layer. This occurs until the output layer is reached.

## E. Ensemble Approach

An ensemble approach uses the help of multiple models are used to predict one outcome. Each model is first trained on a dataset, and then casts a somewhat vote on which target value it predicts. The majority vote of all the models in the ensemble is the final prediction. This voting can be done through two ways: hard and soft voting. Hard voting essentially predicts the class with the most votes at the end from each model. Soft voting predicts the class with the largest summed probability from all the models combined.

## F. Training and Testing

*a) 5 Fold Cross-Validation:* Considering the limited size of the dataset, a 5 fold cross-validation process was chosen to train and test the data. Cross-validation allows for the evaluation of a model by maximizing the amount of training data that can be used upon the model. Through the process of 5 fold cross-validation, the dataset is split into 5 parts (or folds). During the first iteration of the training and testing, the first fold of the dataset is used to test the model while the remainder of the four folds are used to train the dataset. In the second iteration of the training and testing, the second fold of the dataset is used to test data, and the rest of the folds are used to train the dataset. This process repeats using each fold going by each iteration.

## G. Evaluation

*a) Accuracy:* Accuracy is quite a common measure of performance within classification as it is the ratio of correctly predicted observations to the total observations. Accuracy uses the following formula as the basis of its metric:

$$Accuracy = \frac{True\ Positives + True\ Negatives}{All\ Observations}$$

Often, however, accuracy is misleading in the realm of healthcare as it fails to take into account imbalanced datasets. This, however, is not too significant of an issue considering the skewness of the given dataset is quite minimal and hence the imbalance is negligible.[2]

*b) Precision:* Precision is different to accuracy in the sense that it only counts true positives out of all positives. This gives a ratio of all correctly predicted positive observations. A mathematical description of calculating precision is shown below:

$$Precision = \frac{True\ Positives}{True\ Positives + False\ Positives}$$

Precision is a good measure especially with regards to a field where false positives are quite costly. Considering the problem at hand is detecting heart disease, a false positive is not as harmful as a false negative, therefore this benefit does not come into large display with respect to this project. [2]

*c) Recall:* Recall similar to precision predicts true positives, however, ratios them compared to true positives and false negatives. A mathematical interpretation of this ratio is shown below:

$$Recall = \frac{True\ Positives}{True\ Positives + False\ Negatives}$$

This essentially helps measure how accurately the model is able to recognize relevant data.[2]

*d) F1 Measure:* The F1 measure helps combine the efforts of precision and recall by conducting a weighted average. A mathematical representation of this weighted average is shown below:

$$F1\ Score = \frac{2 \times (Recall\ \times Precision)}{Recall + Precision}$$

Therefore, the score takes into account both false positives and false negatives in order to evaluate the model. This is important as the F1 score allows for one to evaluate a model when false negatives and false positives are crucial. Within this problem, a false negative (predicting a patient with a defective heart has a normal heart) may be life threatening, hence an F1 score is required as a false negative is crucial. [2]

*e) AUC-ROC Curve:* The AUC-ROC curve allows for evaluating an algorithm at different thresholds. An ROC is in essence a probability curve which shows the degree of separability. The AUC (Area Under Curve) displays the area under the ROC curve, where a larger AUC indicates that a model is better at predicting a defective heart as defective and a normal heart as normal. [2]

## H. Hyperparameter Tuning

Hyperparameter tuning or hyperparameter optimization is a process in which the optimal parameters are chosen for each model to make the model more efficient at predicting the correct outcome. This process is key in making each model as robust as possible in its predictions. [13]

## V. RESULTS AND EVALUATION

### A. Evalutations of Each Model

Each model, as described in the methodology, was evaluated by five metrics. The evaluation results after being trained and tested through 5 cross validation have been tabulated below in Table 3.

Table.3. Evaluation Scores for Each Model

| Model Name | Accuracy | Precision | Recall | F1 Score | ROC-AUC Score |
|---|---|---|---|---|---|
| Logit | 0.8185 | 0.8848 | 0.8021 | 0.8411 | 0.8238 |
| KNN | 0.6533 | 0.7030 | 0.6771 | 0.6884 | 0.6512 |
| SVM | 0.6533 | **0.9091** | 0.8019 | **0.8513** | 0.8397 |
| Decision Tree | 0.7557 | 0.7939 | 0.7784 | 0.7842 | 0.7692 |
| Gaussian NB | 0.8053 | 0.8242 | 0.8229 | 0.8198 | 0.8096 |
| MLP | 0.7987 | 0.7939 | 0.8310 | 0.8108 | 0.7993 |
| Random Forest | 0.8117 | 0.8606 | 0.8176 | 0.8368 | 0.8229 |
| Soft Voting | **0.8250** | 0.8103 | **0.8848** | 0.8404 | **0.8919** |

The above analysis shows that the Soft Voting Ensemble model was the most robust of all the algorithms, as it had the best score in the majority of employed metrics. Though SVM had better precision and a higher f1 score, the soft voting ensemble model outperformed it on 3 other metrics making it the better predictor

### B. Evaluation on Independent Data

A further evaluation done on the models plotting an ROC-AUC curve is shown below in Figure 5. For this evaluation, a curve was constructed by running the models on independent test data to ensure the model's capability on unseen data sources and to further analyze whether the models were capable of minimizing variance in their prediction. This evaluation proved to show logistic regression as the most powerful model followed by random forest, SVM and the soft voting of all the models. Considering that Logistic Regression did not perform well in the 5 fold cross-validation, yet performed well on the independent testing, one can infer that the great variation between the results may render this instance a coincidence. The ensemble model, SVM, and random forest had minimal variance with regards to an evaluation in both scenarios, hence due to this consistency, one can infer that these models are stable and better at handling the variance factor. The difference between SVM, Random Forest and Ensemble are quite subtle as they lie between 0.01, hence one can suggest that all these models gave a promising result on the independent test data as well.

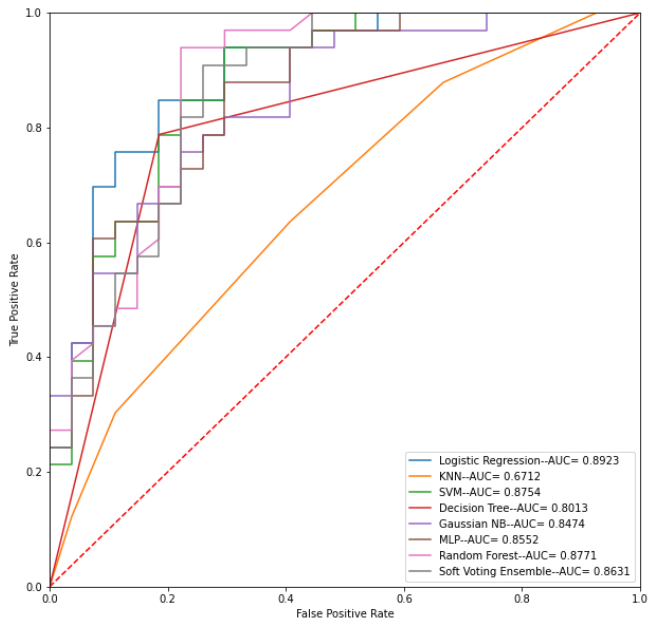

Fig. 5 . ROC-AUC Curve on Independent Test Data

## VI. Conclusion

In conclusion, this project was successful in developing a cardiovascular disease detection algorithm through implementing seven machine learning classification techniques, and further using an ensemble approach to obtain a more robust prediction. Through utilizing attributes from a patient's medical report such as age, sex, resting blood pressure, chest pain, etc. one can predict whether or not a patient could be a victim to cardiovascular disease. Through a rigorous evaluation of the data on both test sets and independent data, it was found that three techniques, being SVM, Random Forest and a soft voting of all seven models, were found to show promising results in predicting heart disease. An ROC-AUC score of 0.8397, 0.8229 and 0.8919 were given to SVM, Random Forest and Soft Voting respectively when considering test data. Though these scores have significant room for refinement, they are enough to aid doctors in predicting a patient's diagnosis quicker, more efficiently and with reduced costs.

It is evident that a clear limiting factor within this project was the size of the database used. As the database only contained 303 rows and only 14 attributes were used, the prediction's quality was limited. Utilizing a larger dataset and using more attributes as predictors one can further refine the algorithm and achieve a prediction more robust in not only correctness, but also in consistency. Furthermore, parameter tuning each model to make it even more suited to the problem would increase the quality of the prediction drastically as during this experiment only a limited number of parameters were optimized with regard to the problem. In the future, this work can be extended through utilizing a larger data set and further employing more algorithms or refining the ones that have already been used in order to achieve the foremost prediction.

Acknowledgment *(Heading 5)*

First of all, I cannot express enough my gratitude towards Mr. Pawel Pratyush as for the completion of this project could not have been accomplished without his guidance and mentorship.

Secondly, I would like to thank Dr. Sushant Kafle, for his guidance and editorial efforts toward my paper.

using Machine Learning, INTERNATIONAL JOURNAL OF ENGINEERING RESEARCH & TECHNOLOGY (IJERT) NCETER – 2021 (Volume 09 – Issue 11).